\def\year{2021}\relax
\documentclass[letterpaper]{article} % DO NOT CHANGE THIS
\usepackage{aaai21}  % DO NOT CHANGE THIS
\usepackage{times}  % DO NOT CHANGE THIS
\usepackage{helvet} % DO NOT CHANGE THIS
\usepackage{courier}  % DO NOT CHANGE THIS
\usepackage[hyphens]{url}  % DO NOT CHANGE THIS
\usepackage{graphicx} % DO NOT CHANGE THIS
\usepackage{booktabs}
\usepackage{amsfonts}
\usepackage{amsmath}
\usepackage{amsthm}
\usepackage{multirow}
\usepackage{subfigure}
\usepackage{threeparttable}  
\usepackage{natbib}  % DO NOT CHANGE THIS AND DO NOT ADD ANY OPTIONS TO IT
\usepackage{caption} % DO NOT CHANGE THIS AND DO NOT ADD ANY OPTIONS TO IT
\nocopyright
\title{Spatio-Temporal Sparsification for General Robust Graph Convolution Networks}
\author{
    Mingming Lu\textsuperscript{\rm 1} Ya Zhang\textsuperscript{\rm 1}\\
    }

\affiliations{
    \textsuperscript{\rm 1}School of Computer Science and Engineering, Central South University\\
    932 Lushan Road, Computer Building\\
    Changsha, Hunan, China 410083\\
    mingminglu@csu.edu.cn

}

\begin{document}

\maketitle

\begin{abstract}
Graph Neural Networks (GNNs) have attracted increasing attention due to its successful applications on various graph-structure data. However, recent studies have shown that adversarial attacks are threatening the functionality of GNNs. Although numerous works have been proposed to defend adversarial attacks from various perspectives, most of them can be robust against the attacks only on specific scenarios. To address this shortage of robust generalization,  we propose to defend the adversarial attacks on GNN through applying the Spatio-Temporal sparsification (called ST-Sparse) on the GNN hidden node representation. ST-Sparse is similar to the Dropout regularization in spirit. Through intensive experiment evaluation with GCN as the target GNN model, we identify the benefits of ST-Sparse as follows: (1) ST-Sparse shows the defense performance improvement in most cases, as it can effectively increase the robust accuracy by up to 6\% improvement; (2) ST-Sparse illustrates its robust generalization capability by integrating with the existing defense methods, similar to the integration of Dropout into various deep learning models as a standard regularization technique; (3) ST-Sparse also shows its ordinary generalization capability on clean datasets, in that ST-SparseGCN (the integration of ST-Sparse and the original GCN) even outperform the original GCN, while the other three representative defense methods are inferior to the original GCN. 
\end{abstract}

\section{Introduction}
Recently, Graph Neural Networks (GNNs) have attracted increasing attention due to its successful applications on various graph-structure data, such as social networks, chemical composition structures, and biological gene proteins~\cite{Zhou2018,Wu2019}. However, recent works~\cite{Sun2018,Xu2019a,Dai2018} have pointed out that GNNs are vulnerable to adversarial attacks, which can crash safety-critical GNN applications, such as auto-driving, medical diagnosis~\cite{Wu2019}. 

To address this issue, numerous works~\cite{Sun2018,Chen2020survey,Jin2020survey} have been proposed to defend the adversarial attacks from the perspectives of data preprocessing~\cite{Wu2019Adversarial}, structure modification~\cite{Wang2019}, adversarial training~\cite{Feng2019}, adversarial detection~\cite{Zhang2019}, and etc.  However, our experimental study and the evaluation in the existing work~\cite{Jin2020survey} have shown that none of the existing defense methods is superior to the others under all attacks for all datasets with all perturbation sizes. This illustrates the limitation of the existing defense methods in terms of the robust generalization capability.

Recently, \cite{Tsipras2019} revealed that the existence of adversarial attack might originate from the utilization of weakly correlated features, which can be reduced by keeping only the strongly correlated features. This phenomenon motivates us to adopt the sparse representation, which is widely utilized in computational neuroscience~\cite{Ahmad2019}, to reduce the effect from the weakly correlated features. 
Thus, in this work, we
propose a spatio-temporal feature sparsification framework to improve the robustness of the GNN models. 

The spatial feature sparsification (called TopK) in the proposed framework simply keeps the $k$ features with the largest values and sets all the other features to zero. In spirit, TopK is the same as the Dropout regulirazation technique~\cite{srivastava2014dropout} except that Dropout randomly drops neurons, while TopK orders the neurons according to their output values and keeps only the neurons with the top $k$ values. Through experiment studies, we identify that TopK can improve the defense performance under four representative adversarial attacks on three typical benchmark datasets with varies perturbation sizes. 

However, the robustness brought by TopK is at the expense of the generalization capability. Compared with Dropout, TopK loses the randomness, which sacrifices the generalization capability, as the randomness in Dropout can decompose a complex model into an ensemble of a large number of simpler models. To address this issue, temporal feature sparsification is introduced to alternate the non-zero features (also called active features) in each training epoch. Through the feature alternation, more features can participate in node representation in turn. Thus, the spatial sparsification together with the temporal sparsification (abbreviated as ST-Sparse) can behave similarly to the Dropout regulirazation technique. Therefore, ST-Sparse might achieve the similar generalization capability as Dropout. Moreover, through experiment evaluation, we identify that ST-Sparse can achieve robust generalization in that it can integrate with the existing defense methods to further improve the model robustness, similar to the
integration of Dropout into various deep learning models as
a standard regularization technique.

\begin{figure}[ht]
	\centering
	\includegraphics[width=\linewidth]{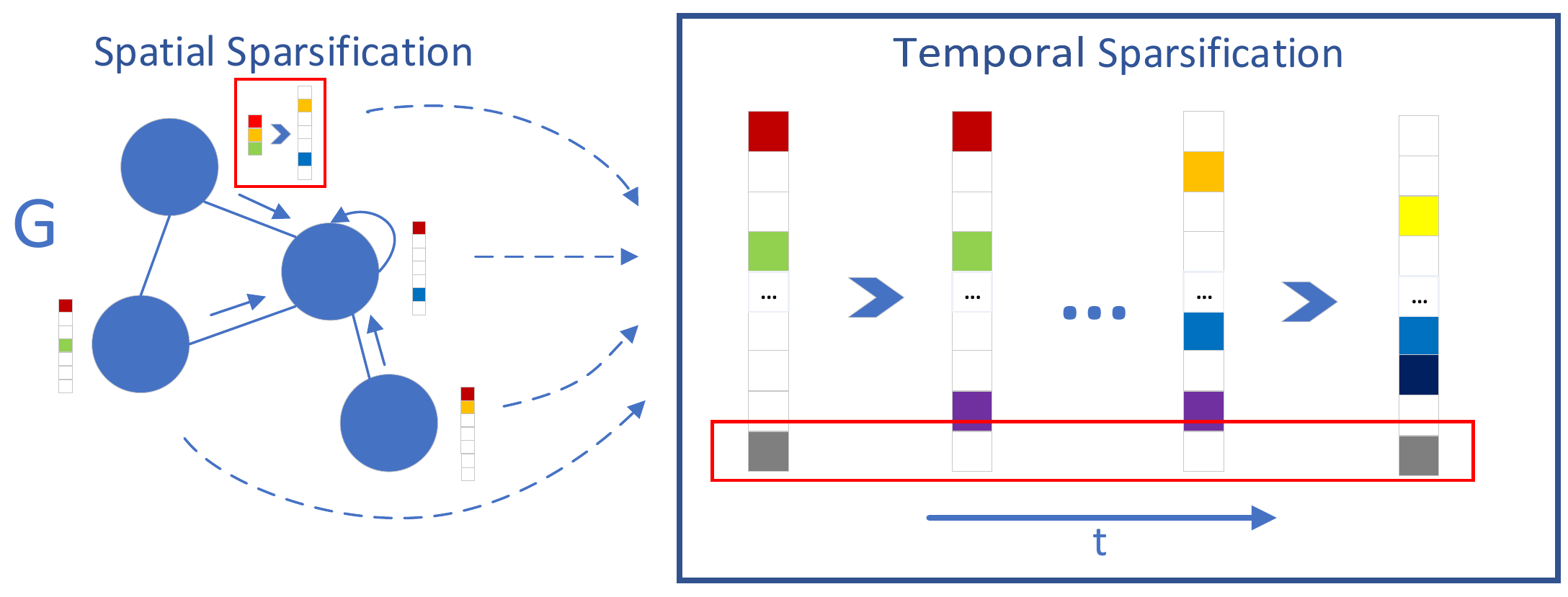}
	\caption{The illustration of spatio-temporal sparsification, where the vertial rectangular bar associated with each node represents the node's feature vector and the colored/white squares in the bar denote active/inactive features. In the temporal sparsification part, the horizontal red rectangle illustrates the on-and-off activation patten of temporal sparsity.}
	\label{fig:ST-sparseGCN_illustration}
\end{figure}

Fig.~\ref{fig:ST-sparseGCN_illustration} illustrates the basic idea of the proposed ST-Sparse mechanism. The spatial sparsification is mainly dedicated to transform a dense hidden node vector of a GNN to a sparse high-dimensional vector, where only the top $k$ salient features are activated, as illustrated through the red rectangle at the top left part of Fig.~\ref{fig:ST-sparseGCN_illustration}. 
The temporal sparsification further sparsifies the active features along the time dimension during the GNN training process. More specifically, the duty cycle of each active feature dimension is sparse so that each active feature will not be intensively used. 

Note that the temporal sparsification is applied to the feature dimension instead of the features of individual nodes, because on one hand, the salient features of individual nodes usually focus only on a few dimensions, the temporal spasification of these features may significantly degrade the model performance; on the other hand, the overall distribution composed of all nodes can better reflect the temporal sparsity. By balancing the duty cycle of activation among different dimension, it is possible to avoid the intensive usage of certain dimensions, thereby increasing the model's expressive capability, which in turn increases the robustness of the model.

\iffalse 
Furthermore, although the adversarial attack can be applied to GNN either through edges or node features, the correponding perturbation can be reflected through the aggregated node features. Thus, both the spatial and temporal sparsification are applied to the aggregated features. 
\fi
The main contributions of this work is summarized as follows.
\begin{itemize}
	\item From the perspective of spatio-temporal sparsity, we explore to construct a robust feature space, where the information propagation in GNN is less vulnerable to adversarial attacks.
	\item We provide a novel ST-Sparse mechanism, which utilizes TopK to realize spatial sparsity in high-dimensional vector space, and adopts attention to balance the activation duty cycles among different dimensions, so as to realize the temporal sparsity in the feature space. 
	\item To verify the effectiveness of ST-Sparse, we apply ST-Sparse to the graph convolution network (GCN)~\cite{Kipf2019} (denoted as ST-SparseGCN). Intensive experiments through three benchmark datasets show that ST-SparseGCN can significantly improve the robustness, robust generalization, and ordinary generalization of GCN in terms of classification accuracy. 
\end{itemize}

%The benefits of feature sparsification can be summarized as follows: (1) it is a novel design perspective for denfense models on graph adversarial attacks

\section{Related Works}\label{sec:related}
\iffalse most of the exisiting GNN defense methods either assumed certian prior knowledge on the graph structures or required time-consuming optimization techniques for training a robust GNN model. For example, \cite{Zhu2019} assumed the existence of the correlation between the perturbations incurred by the attacks and the GNN variances,  
\cite{Wu2019Adversarial} assumed the Jaccard similarity among neighbor nodes can well characterize the perturbations incurred by the attackers, and adversarial training~\cite{Feng2019} needs to solve a complex min-max optimization problem.\fi

{\bf Adversarial attacks on general graph.} The basic idea of adversarial attacks on graph is to change the graph topology or feature information to intentionally interfere with the classifier. \cite{Dai2018} studied a non-target evasion attack based on reinforcement learning. \cite{Zugner2018} proposed netattack, a  poisoning attack on GCN, which modifies the training data to misclassify the target node. Further, ~\cite{Zugner2019} used the meta-gradient to solve the min-max problem in attacks during training, and proposed an attack method that reduces the overall classification performance. Besides, \cite{Xu2019} simplified the discrete graph problem by convex relaxation, and thus proposed a gradient-based topological attack.

{\bf Defense methods on general graph.} The existing defense methods can be classified from the perspectives of data preprocessing~\cite{Wu2019Adversarial}, structure modification~\cite{Wang2019}, adversarial training~\cite{Feng2019}, the modificaiton of the objective function~\cite{NIPS2019_9041}, adversarial detection~\cite{Zhang2019}, and hybrid defense~\cite{DBLP:journals/corr/abs-1903-05994}. The proposed ST-SparseGCN model can be regarded as a structure modification methods, because it modifies the original GCN structure, as shown in Fig.~\ref{fig:ST-sparseGCN_framework}. However, the proposed ST-Sparse defense methods can also be integrated with the other GCN defense models, such as 
GCN-Jaccard\cite{Wu2019Adversarial} and GCN-SVD\cite{Entezari2020}, which can be regarded as data preprocessing methods. Thus, the integrated models can be classified as hybrid defense models. Although dozens of defense methods on graph have been proposed, none of them  shows the robust generalization, as they are not superior to the others under all attacks for all datasets with all perturbation sizes~\cite{Jin2020survey}. 

\iffalse \cite{Wu2019Adversarial} improved the model robustness by comparing the Jaccard similarity between a node and its neighbor nodes, and then removing those neighbors with low similarity. This method is essentially an adversarial example detection scheme, which can be used as a constraint to adversarial attacks. Since adversarial attacks come from the feature/edge perturbation on the graph, \cite{Zhu2019} modeled the hidden representations of each node as Gaussian distribution
and proposed to reduce the allowable perturbation from adversarial attacks through 
reducing the distribution variance so that the impact of adversarial attacks will also be reduced. \cite{Zugner2019a} proposed a robust certificate based on convex relaxation, which considers only  the perturbation on node featuress, and used the semi-supervised property to improve the robustness of the model. \cite{Tang2019} learned the ability to punish perturbations through the information of additional clean graphs from similar domains and transfered them to the target poisoning graph.\fi

{\bf Sparsity and Robustness.} The relation between sparsity and robustness has been revealed in the fields of image classification~\cite{Guo2018} and neuroscience~\cite{Ahmad2019}. From the perspective of image classification, \cite{Guo2018} clarified the inherent relation between sparsity and robustness through theoretical analysis and experimental evaluation.
\cite{Cosentino2019,Tsipras2019} revealed that the existence of adversarial
attack might originate from the utilization of weakly
correlated features, which can be reduced by keeping only
the strongly correlated features. This phenomenon also illustrated the necessity
of sparsity to reduce the effect of the weakly correlated features.

{\bf Difference to the Existing Methods.} Unlike the existing works on GNN robustness, most of which assume certain prior knowledge concering the attack, we intend to construct a robust feature space that can resist attack without any prior knowledge on attack, which can be called ``black box defense". \cite{ZhengICML20} also considered the relation between the model robustness and sparsity. However, its sparsity is defined for the sparsity of the graph structure, instead of the sparsity of the hidden node representation introduced by our ST-Sparse method. It is worth to note that in ST-Sparse, since the perturbation is injected in the hidden layer, it does not have to generate perturbation on graph structure and node feature separately.  
\section{Preliminaries}\label{sec:pre}

\subsection{Notations}\label{sec:notations}

Given an undirected graph $G=(V,E,X)$, where $V=\left\{v_1,v_2,...,v_n\right\}$ is a set of nodes with $|V|=n$, $E \subseteq V\times V$ is a set of edges that can be represented as an adjacency matrix $A\in {{\left\{ 0,1 \right\}}^{n\times n}}$, and $X=\left[ {{x}_{1}},{{x}_{2}},\ldots , {{x}_{n}} \right]^T \in {{\mathbb{R}}^{n \times d}}$ is a feature matrix with $x_i$ denoting a feature vector of node $v_i \in V$. $\text{C}=\left< {{c}_{1}},{{c}_{2}},\ldots ,{{c}_{n}} \right>$ is the class label vector with $c_i$ representing the label value of node $v_i$. 

\iffalse 
We define $h_i=\left<h_{i1},h_{i2},\ldots ,h_{id} \right>\in \mathbb{R}^{d}$  as a hidden representation of node $v_i$ with $h_{ij}$ denoting its $j$-th feature and $d$ representing the dimension of its feature space. Meanwhile, we use  $h'_i=\left<h'_{i1},h'_{i2},\ldots ,h'_{id_h} \right>\in \mathbb{R}^{d_h}$ the high dimensional verson of $h_i$, where $d \ll d_h$. $d_h$ represents the dimension of the high-dimensional feature space. 
\fi

\subsection{Graph Convolution Networks}
In the paper, we focus on GCNs for node classification. In particular, we will consider the well established work \cite{Kipf2019}. As a semi-supervised model, GCN can learn the hidden representation of each node. The hidden vectors of all nodes in the $l+1$ layer can be represented recursively by the hidden vectors of the $l$ layer as follows.

\begin{equation}\label{eq:GCN}
{{H}^{\left( l+1 \right)}}=\sigma \left( {{{\tilde{D}}}^{-\frac{1}{2}}}\tilde{A}{{{\tilde{D}}}^{-\frac{1}{2}}}{{H}^{\left( l \right)}}{{W}^{\left( l \right)}} \right)
\end{equation}
where ${\tilde{A}}={A}+{{I}_{n}}$, $W^{\left( l \right)}\in {\mathbb{R}}^{d^{(l)}\times d^{(l+1)}}$yl denotes the learnable weight matrix at layer $l$, ${{\tilde{D}}_{i}}=\mathop{\sum }_{j}{{\tilde{A}}_{ij}}$, and $\sigma\left( \cdot  \right)$ is an activation function, such as ReLu. Initially,  $H^{(0)}=X$.

\section{The ST-SparseGCN Framework}\label{sec:model}
\begin{figure*}[ht]
	\centering
	\includegraphics[width=0.7\linewidth]{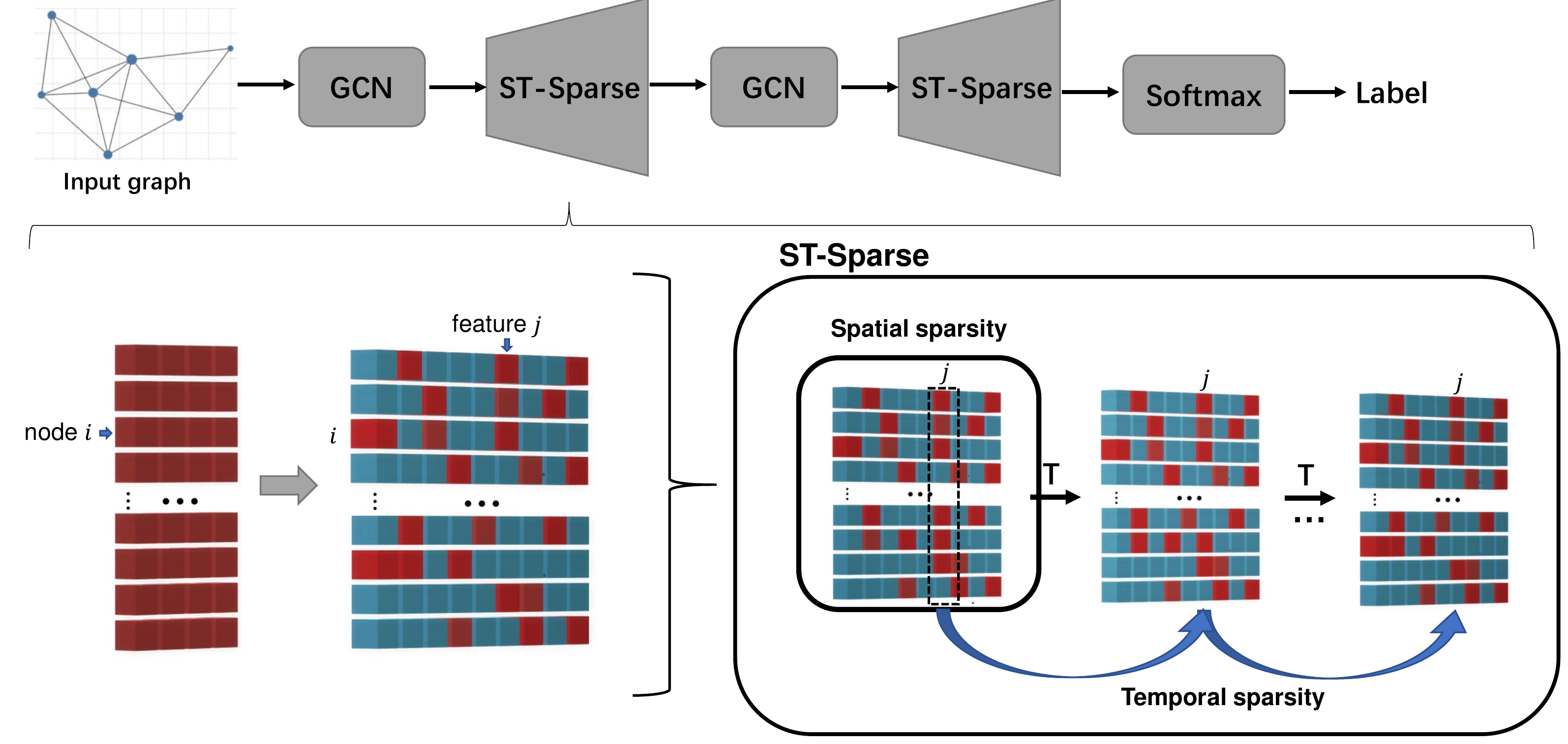}
	\caption{The SparseGCN framework.}
	\label{fig:ST-sparseGCN_framework}
\end{figure*}

In the following, we introduce technical details of the proposed ST-SparseGCN. As shown in Fig.~\ref{fig:ST-sparseGCN_framework}, ST-Sparse can be integrated into the GCN model as an activation layer through replacing the ReLu activation function. The ST-Sparse layer will transform the dense feature $h_i$ of each node $v_i$ into a ST-Sparse feature $s_i$. While the ST-Sparse feature transforming process can be further decompose into the spatial sparsification and temporal sparsification processes. 

%\subsection{The SparseGCN Model}

\iffalse
the ST-Sparse layer transforms $H$, the dense representation output by the GCN layer, into a high-dimensional sparse represenation $S$. 

And $S$ follow the learning rules based on Spatio-Temporal Sparse, that is, activates the base features with large feature values at each epoch, and assigns attention weights to the activation frequencies of different base features in the high-dimensional sparse representation during the training.
\fi

\subsection{The High-dimensional Sparse Space}\label{sec:highDimension}
First, we will describe the mapping from the dense space to the high-dimensional space, which can be simply realized through replacing the parameter matrix $W^{(l)}$ in Eq.  (\ref{eq:GCN}) with a high-dimensional version $W^{(l)}_h$, as shown in Eq.  (\ref{eq:highGCN}).
\begin{equation}\label{eq:highGCN}
{H^{\left( {l + 1} \right)} = \sigma \left( {{{\tilde D}^{ - \frac{1}{2}}}\tilde A{{\tilde D}^{ - \frac{1}{2}}}H^{\left( l \right)}W_h^{\left( l \right)}} \right)},
\end{equation}
where $W^{\left( l \right)}\in {\mathbb{R}}^{d^{(l)}\times d_h}$.
Compared to $d^{(l+1)}$ (the second dimension of $W^{(l)}$ in Eq. (\ref{eq:GCN})), $d_h$ (the second dimension of $W^{(l)}_h$) is much larger. In Section \ref{sec:para}, we will illustrate the underlying reason for high dimensional space through experiment evaluation, which will show that the low dimension can significantly reduce the performance of the proposed ST-SparseGCN. Thus, the high-dimensional space is one of the key factors for the effectiveness of the propose ST-SparseGCN. It is worth to note that $d_h$ is the same for all layers except the input layer, i.e., $\forall l\ge 1$, $H^{(l)}\in {\mathbb{R}}^{n\times d_h}$ and $H^{(0)}=X\in{\mathbb{R}}^{n\times d}$.

Next, we will formally introduce the definition of spatial sparsity as follows. 
\newtheorem{myDef}{Definition}
\begin{myDef}\label{def:spatialSparsity}
	{Spatial Sparsity.}
	$\forall v_i\in V$, its high-dimensional feature vector $h_i=<h_{i1},h_{i2},\ldots ,h_{id_h}>$ satisfies the spatial sparsity if $||h_i||_0 \ll d_h$, where $||\cdot||_0$ denote the $l_0$-norm, i.e. the number of non-zero elements.
\end{myDef}
Def.~\ref{def:spatialSparsity} implies that the non-zero elements of a spatial sparse vector should be much less than the vector dimension. In the following, we will adopt $s_i$ to denote the sparse version of $h_i$. Also, $S=[s_1,s_2, \ldots ,s_n]^T$ represents the sparse matrix consisting of the sparse vectors from all nodes.

Although spatial sparsity can ensure the feature sparsity of individual nodes, it cannot  guarantee that individual features are sparse, i.e., the number of nodes activated on any given feature is much less than the total number of nodes. For example, in Fig~\ref{fig:ST-sparseGCN_framework}, feature $j$ is not temporally sparsed after spatial sparsification because too many nodes activate feature $j$. Through temporal sparsification, the non-zero elements associated with feature $j$ will be gradually reduced.   

This new type of sparsity can be illustrated through simple calculation. If $\forall t \in \left[ {1,2, \ldots ,{\rm{T}}} \right]$, where $T$ is the total number training epochs, and $\forall v_i\in V$, $||s^t_i||_0\le k$, then $||S^t||_0\le n\times k$,  where $s^t_i$ and $S^t$ represent $s_i$ and $S$ at epoch $t$, respectively, as there are $n$ nodes in total. Thus, on average, each feature will be on duty (i.e., non-zero values) for at most $\frac{n\times k}{d_h}$ nodes, because there are $d_h$ features in total. Since $k\ll d_h$ according to Def.~\ref{def:spatialSparsity}, it can be concluded that $\frac{n\times k}{d_h}\ll n$, where $n$ is actually the maximal number of non-zero elements for any feature at epoch $t$. Thus, from the feature's perspective, if the duty cycle (in terms of non-zero elements) for all features needs to be balanced, it also shows the sparsity phenomenon.  

The underlying reason for the necessity of the duty-cycle balance lies in that, if a feature is on duty for too many nodes, this feature may show the Mathew effect, i.e., the more a feature is used at the current epoch, the more oftern it will be used in the following epochs. This Mathew effect can be took advantaged by the adversarial attacker through manipulating the heavily used feature. 

Thus, it is desirable to introduce temporal spasity so that the duty cycle of features can be balanced along with the training epochs. To formally define temporal sparsity, we introduce $s_{*j}^t$ to denote the set of nodes that utilizes the $j$-th feature, which is equal to the $j$-th column of $S^t$, i.e., $s_{*j}^t=<s_{1j}^t,s_{2j}^t,\ldots ,s_{nj}^t>$.

Based on the above description, the temporal sparsity concerning the $j$-th feature  can be formally defined as follows.
\begin{myDef}\label{def:temporalSparsity}{Temporal Sparsity.}
	For $\forall j\in \{1,\cdots, d_h\}$, the vector $<s_{*j}^1,\ldots, s_{*j}^t, \ldots ,s_{*j}^T>$ satisfies the temporal sparsity if $$\lim_{T\to +\infty }\frac{sum_{t=1}^T ||s_{*j}^t||_0}{T} = \frac{n\times k}{d_h} $$
\end{myDef}

\subsection{TopK Based Spatial Sparsification}

In our ST-SparseGCN model, the spatial sparsification is implemented through TopK, which simply selects the top $k_{\alpha}=\lfloor \alpha \cdot d_h\rfloor$ features for any $h_i\in H$, where $\alpha\in \left(0,1\right)$ is the spatial sparse ratio. TopK can be formalized as follows. 
\iffalse
\begin{equation}\label{eq:topK}
z_i = TopK({h_i,k_{\alpha}}),
\end{equation}
\fi

\begin{equation}\label{eq:topK2}
s_i=TopK({h_i,k_{\alpha}})= \left\{ {\begin{array}{*{20}{c}}
		{h_{ij},\qquad j \in {z_i}},\\
		{0,\qquad \;\;\; j \notin {z_i}.}
		\end{array}} \right.
\end{equation}
where $z_i$ represents the set of features with the largest $k_{\alpha}$ values from $h_i$. In another words, TopK will keep the values of those top $k$ features and set all the other features to zero.

Through replacing the activation function in Eq.(\ref{eq:highGCN}) with TopK, we can implement a spatial sparsed GCN, which can be formalized through the equation shown in Eq.(\ref{eq:spatialSparseGCN}).

\begin{equation}\label{eq:spatialSparseGCN}
S^{(l + 1)} = TopK ({\tilde D}^{-\frac{1}{2}}{\tilde A}{\tilde D}^{-\frac{1}{2}}S^{(l)}W_h^{(l)}, k_{\alpha}),
\end{equation}
where $S^{(0)}=X$ initially. It is worth to note that the TopK function in Eq.(\ref{eq:spatialSparseGCN}) is a matrix version of the TopK fuction in Eq.(\ref{eq:topK2}). This matrix version selects the top $k_{\alpha}$ features for each node $v_i$ independently.    

The spatial sparse ratio $\alpha$ in the TopK function is a hyperparameter to be adjusted. Intuitively, on  one hand, small $\alpha$ implies less non-zero features, which might seriously compromise the generalization capability of the proposed model, because the possible vectors that can be represented in the high-dimensional space will become less along with smaller $k_{\alpha}$. On the other hand, large $\alpha$ may compromise the model robustness. The appropriate value of $\alpha$ will be evaluated in Section \ref{sec:perturb}.
%is usually set between 0.1 and 0.2 in experiments

\begin{figure}[ht]
	\centering
	\subfigure[The comparison of activated neurons.]{
		\includegraphics[width=0.47\linewidth]{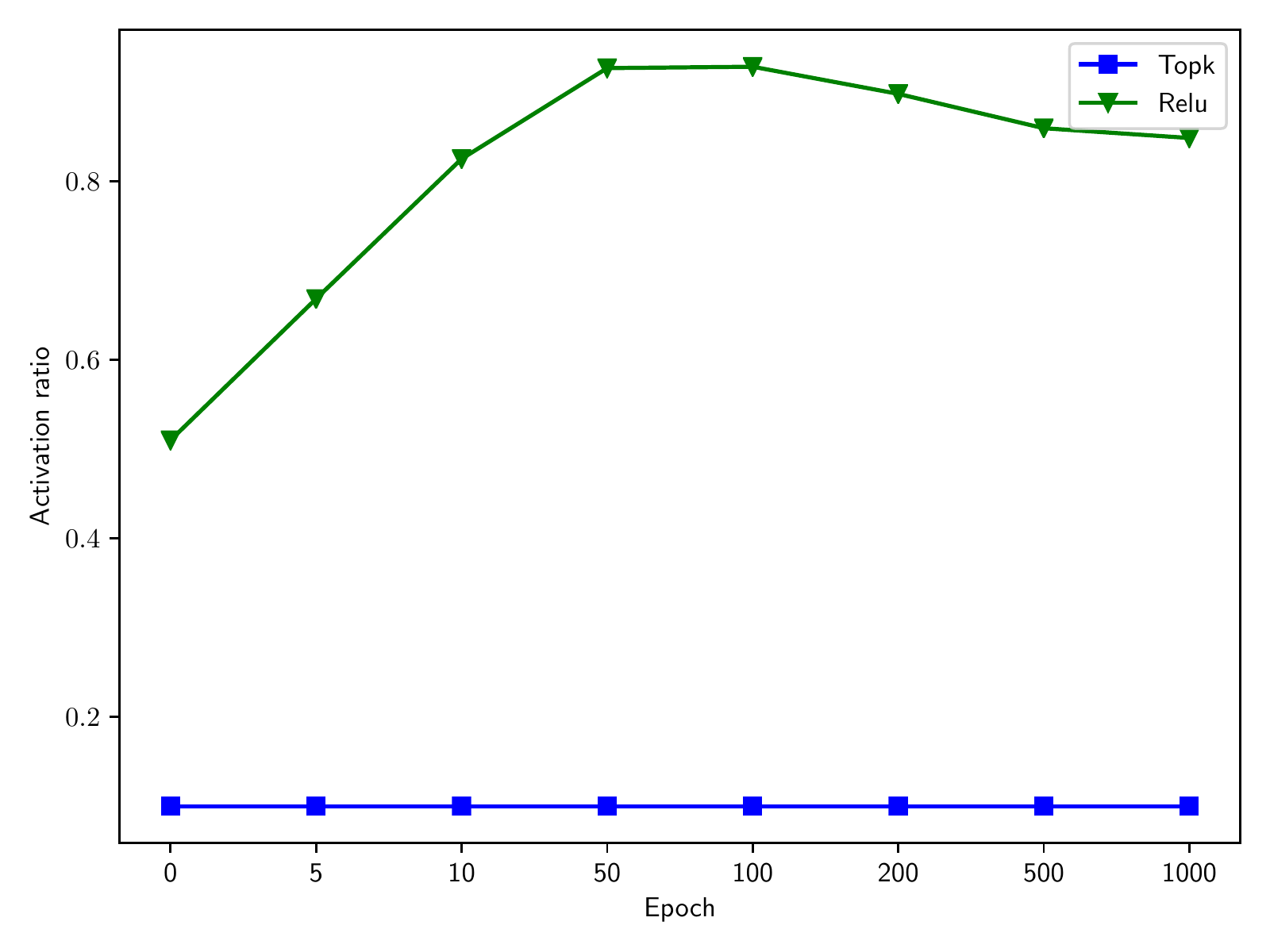}
		\label{fig:rt:a}
	}
	\subfigure[The comparison of function curves.]{
		\includegraphics[width=0.47\linewidth]{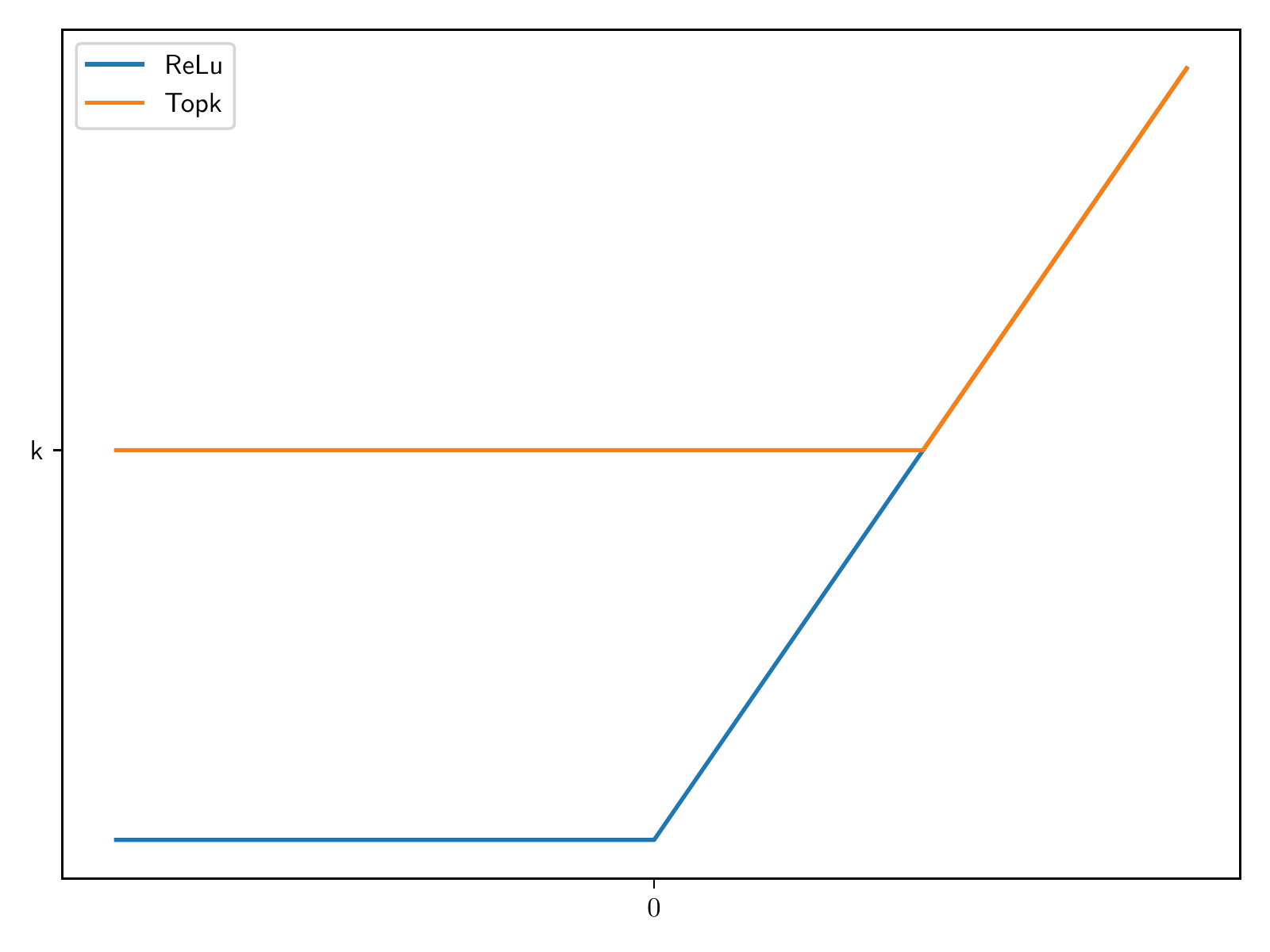}
		\label{fig:rt:b}	
	}
	\caption{The comparison of TopK and ReLu.}
	\label{fig:rt}
\end{figure}

{\bf TopK VS. ReLu.} In ST-SparseGCN, the ReLu function in GCN has been replaced by the TopK function. The effect of the replacement can be illustrated through Fig. \ref{fig:rt:a}, where the GCN coupled with TopK and the GCN with ReLu are compared in terms of the ratio of activated neurons during the training process. From Fig. \ref{fig:rt:a}, it can be observed that TopK greatly reduces the number of activated neurons. 
TopK and ReLu can also be compared through the funciton curves as shown in Fig. \ref{fig:rt:b}, from which it can be observed that it is more difficult for a neuron to be activated through the TopK activation function. 

%This means that TopK activation has the full power of ReLu.

{\bf The Cost of Robustness.} \cite{xiao2020enhancing} has proved that the computational complexiy of TopK is asymptotically $O(N)$, which is the same as ReLu. However, it takes more time for TopK to converge in experiments, which might originate from the spatial sparsity. Due to the spatial sparsity, only a small number of neurons can be activated, which implies that the gradient update covers only a small number of neurons in each epoch. Nevertheless, the computing cost of TopK can be reduced through the computing optimization of sparse matrix.

\subsection{Attention Based Temporal Sparsification}
\iffalse
Spatial sparsity can utilize the salient base features to well characterize the property of the corresponding nodes. However, spatial sparsity may incur the Mathew effect: the salient features  become more salient, while the unimportant features become even less important. This Matheew effect may significantly compromise the generalization capability of the spatial sparsified GCN, because the possible vectors that can be represented in the high-dimensional space will be significantly reduced.
\fi

At the first glance, it seems that temporal spasity can be realized through applying the TopK function as follows:  $TopK(s^t_{*j},\frac{n\times k}{d_h})$. However, this may reduce the number of non-zero features for certain nodes, which might compromise the generalization capability as discussed previously. Furthermore, it may cause a sudden discontinuities in the model output. 

To avoid the above issues, we propose an attention-based temporal sparsification mechanism, where at any epoch $t$, for any node $v_i$, its feature $j$ is assigned an attention value $b_{ij}^t$.  This attention value will be adaptively adjusted according to the historical sparsity information of feature $j$, namely, $||s^{t'}_{*j}||_0$ ($t'\in \{1,2,\cdots, t-1\}$). Then, the adjusted attention value $b_{ij}^t$ is used as a weight to adjust the corresponding feature $j$ of node $v_i$ in the spatial sparsified GCN hidden representation (namely $S^{(l)}$, as defined in Eq.(\ref{eq:spatialSparseGCN})) so that the feature with larger sparsity value will reduce its chance to be selected by the TopK function. 

Concretely, in each epoch $t$, the attention mechanism updates the attention value of each node $v_i$'s feature $j$ based on the integration of the historical sparsity $\hat{s}_{ij}^t$ and the current sparsity of the feature (i.e. $||s^t_{*j}||_0$). If integrated feature sparsity  is higher than the sparsities of the other features, its attention value $b^t_{ij}$ will be reduced accordingly.   

Formally, the integrated sparsity of any feature $j$ associated with node $v_i$ is updated as  shown in Eq.(\ref{eq:sparsityIntegration}). 
 \begin{equation}
\label{eq:sparsityIntegration}
{\hat{s}_{ij}^{t + 1} = \hat{s}_{ij}^t + \tau \times ||s^t_{*j}||_0},
\end{equation}where $\hat{s}_{ij}^t$ is the historical sparsity of node $v_i$'s feature $j$ before epoch $t$ and $\tau$ is a hyper parameter that controls the decay rate of historical information. Initially, $\hat{s}_{ij}^0 = 0$. 

Based on the integrated sparsity $\hat{s}_j^t$, the attention value $b^t_j$ is updated through a smooth exponential function as shown in Eq.(\ref{eq:attentionUpdate})
\begin{equation}\label{eq:attentionUpdate}
b_{ij}^t = \exp(-\gamma\hat{s}_{ij}^t),
\end{equation}where $\gamma$ is a hyper parameter. From Eq.(\ref{eq:attentionUpdate}), it can be observed that, the larger the integrated sparsity $\hat{s}_{ij}^t$, the smaller the updated attention value $b_{ij}^t$, because the smooth exponential function $\exp(\cdot)$ is a monotonically decreasing function. 

From Eq.(~\ref{eq:sparsityIntegration}), it can be observed that the historical sparsity $\hat{s}_{ij}^t$ is actually independent of nodes, and so does the attention value $b^t_{ij}$, according to Eq.(\ref{eq:attentionUpdate}). Thus, Eq.(~\ref{eq:sparsityIntegration}) and Eq.(\ref{eq:attentionUpdate}) can be computed only once for all nodes, from which we can obtain feature $j$'s historical sparsity and attention values for all nodes, namely vector $\hat{s}_{*j}^t$ and vector $b^t_{*j}$, respectively.

From $b^t_{*j}$, $j\in\{1,\cdots, d_h\}$, we can contruct the attention mask matrix ${\cal B}^t$ as follows. 
\begin{equation}\label{eq:attentionMatrix}
{\cal B}^t = <b_{*1}^t, \ldots,b_{*j}^t,\ldots,b_{*d_h}^t>,
\end{equation}

\iffalse
It is worth to note that $\hat{s}^t$  is shared among all nodes in the graph, because temporal sparsity focuses on feature space instead of  the features of individual nodes.
 Thus, the estimated activation frequency  $\hat{S}^t \in \mathbb{R}^{n \times d_h}$ can be formalized as follows.
\begin{equation}
\hat{S}^t = <\hat{s}^t, \ldots, \hat{s}^t, \ldots, \hat{s}^t>,
\end{equation}

To reduce the Mathew effect, a smooth exponential function shown in  Eq.~(\ref{eq:exponential}) is introduced to adjust the attention weight of each feature dimension, so that the weight of the feature with high/low activation frequency will be reduced/improved accordingly.  
\begin{equation}\label{eq:exponential}
{{{\cal B}_t} = \exp (-\gamma {B^t})}
\end{equation}
\fi

Based on the attention mask matrix, the proposed ST-SparseGCN can be formalized through  Eq. (\ref{eq:ST-SparseGcN}).
\begin{equation}\label{eq:ST-SparseGcN}
\begin{split}
S^{(l+1)}=TopK(&\tilde{D}^{-\frac{1}{2}} \tilde{A} \tilde{D}^{-\frac{1}{2}}\left(\mathcal{B}_{t}^{(l)} \odot S^{(l)}\right) W_{h}^{(l)},k_{\alpha}).
\end{split}
\end{equation}
Initially, $\mathcal{B}_t^{(0)}=0 $ and $S^{(0)}=X$. Eq. (\ref{eq:ST-SparseGcN}) can be desribed as follows. At epoch $t$, in the $(l+1)$-th layer, sparse matrix $S^{(l)}$ first  multiplies the attention mask matrix $\mathcal{B}_{t}^{(l)}$ element-wisely to temporal sparsify the feature space, so as to mitigate the Mathew effect. The sparsified matrix will be fed as input into GCN for information propagation among nodes. The GCN output is further spatially sparsified thourgh TopK activation fucntion. In the end, $S^{(L)}$ is passed to a fully connected layer with the softmax activation function to predict labels $Y$.

\section{Experimental Evaluation}
\label{sec:experiment}
\subsection{Experimental Settings}
{\bf Baselines.}
To evaluate the robustness and effectiveness of ST-SparseGCN, experiments are performed on the deep learning framework PyTorch~\cite{Steiner2019} and the GNN extension library PyG~\cite{Fey2019}. The proposed defense model (ST-SparseGCN) is compared with four baselines, three of which are representative graph defense methods, on the task of node-level semi-supervised classification as follows.
\begin{itemize}
\item GCN\cite{Kipf2019}:GCN proposes to simplify the graph convolution using only the first order polynomial, i.e. the immediate neighborhood. By stacking multiple convolutional layers, GCN achieved the state-of- the-art performance in clean datasets.
\item GCN-Jaccard\cite{Wu2019Adversarial}: GCN-Jaccard utilizes the Jaccard similarity of features to prune perturbed graphs based on the assumption that the connected nodes usually show high feature similarity.
\item GCN-SVD\cite{Entezari2020}: GCN-SVD proposes a low-rank representation method, which can  approximate the original node representation with a low-rank representation, so as to resist the adversarial attacks.
\item RGCN\cite{Zhu2019}: RGCN aims to defend against adversarial edges with Gaussian distributions as the latent node representation in hidden layers to absorb the negative effects of adversarial edges.
\end{itemize}
We implement the above baseline methods with refering to the implementation in DeepRobust\cite{li2020deeprobust}.

{\bf Attacker models.}
To validate the defensive ability of our proposed defender, we choose four representative GCN attacker models.
\begin{itemize}
	\item DICE\cite{Waniek2018}:DICE randomly selects node pairs to flip their connectivity (i.e., the removal of the existing edges and the connection of non-adjacent nodes).
	\item Mettack\cite{Zugner2019}: Mettack aims at reducing the overall performance of GNNs via meta learning. We used the attack method Meta-Self.
	\item PGD\cite{Xu2019}:PGD is projected gradient descent topology attack to attacking a pre-defined GNN.
	\item Min-Max\cite{Xu2019}: Min-max is min-max topology attack to attacking a re-trainable GNN. The minimization is optimized using the PGD method and the maximization aims to constrain the attack loss by retraining $W$.
	
\end{itemize}

{\bf Parameter Setting.}
The following common parameters are the same for ST-SparseGCN and the baselines. The number of GCN layer is 2 and the training epochs is 200. The selected optimizer is  Adam~\cite{Kingma2015} with a fixed learning rate of 0.01. The other hyperparameters in baselines are closely followed the benchmark setup. And the hyper parameters in the ST-SparseGCN model are adjusted based on the validation set to achieve the best robust performance. Parameter sensitivity of ST-SparseGCN will be analyzed in Section \ref{sec:para}. The final results of all experiments are obtained by averaging 5 repeated experiments. Our experiments are performed on NVIDIA RTX 2080Ti GPU.

{\bf Datasets.}
ST-SparseGCN is evaluated on three well-known datasets: Cora, Citeseer and Polblogs~\cite{Sen2008}, where nodes represent documents and edges represent citations. The sparse bag-of-words feature vector associated with each node is the model input. Table \ref{tab:dataset} enumerates the statistics information of the datasets.  The same training set, test set, and validation set from the same data set is used to fairly evaluate the performance of different models.

\begin{table}	
	\caption{Statistics of datasets}
	\centering
	\begin{tabular}{lllll}
		\toprule
		& Nodes & Edges & Features & Classes \\
		\midrule
		Cora       & 2708(1 graph) & 5429 & 1433 & 7      \\
		Citeseer   & 3327(1 graph) & 4732 & 3703 & 6    \\
		Polblogs     & 1490(1 graph) & 33430 & 1490 & 2  \\
		\bottomrule
	\end{tabular}
	
	\label{tab:dataset}
\end{table}

\subsection{Classification Performance Evaluation}
In order to properly measure the impact of the perturbation, we first evaluated the performance of ST-SparseGCN and all baselines on different clean datasets. The average accuracy with standard deviation is enumerated in Table \ref{tab:clean}, which indicates that  ST-SparseGCN can achieve excellent performance on clean data sets. Compared to four baselines, the superiority of ST-SparseGCN may come from the generalization capability of the temporal sparsification on the feature space. 

\begin{table}[ht]
	\caption{The results of accuracys($\%$) on clean datasets}
	\centering
	\begin{tabular}{llll}
		\toprule
		& Cora     & Citeseer & Polblogs    \\ \midrule
		GCN          & 81.6$\pm$0.6 & 70.7$\pm$0.8 & 85.9$\pm$0.9  \\
		GCN-Jaccard  & 78.9$\pm$0.8 & 71.4$\pm$0.7 & 50.4$\pm$0.9 	\\
		GCN-SVD      & 68.4$\pm$0.8 & 59.8$\pm$0.9 & 80.4$\pm$0.4 \\
		RGCN         & 81.1$\pm$0.6 & 71.4$\pm$0.5 & 85.3$\pm$0.7   \\
		ST-SparseGCN & \textbf{82.2$\pm$0.6} & \textbf{72.0$\pm$0.6} & \textbf{89.1$\pm$0.4}   \\ 
		\bottomrule
	\end{tabular}
	
	\label{tab:clean}
\end{table}

\begin{table*}[ht]
	\caption{Summary of \textbf{mDR}s(in percent) in classification accuracy compared to GCN  in the clean/original graph.Lower is better.}
	\centering
	\begin{tabular}{lllllllllllllll}
		\toprule
		Dataset              & \multicolumn{4}{c}{Cora}                                        &  & \multicolumn{4}{c}{Citeseer}                                   \\  \cmidrule{2-5} \cmidrule{7-10} 
		Defender \textbf{/} Attacker  & DICE          & Mettack        & MinMax       & PGD           &  & DICE          & Mettack        & MinMax      & PGD               \\ \midrule
		GCN                  & 5.28          & 54.13          & 21.90          & 10.09         &  & 2.53          & 64.39          & 22.82         & 5.74          \\
		GCN\_Jaccard         & 6.82          & 38.51          & 13.18          & 17.57         &  & 1.12          & 53.08          & 12.16         & 4.13          \\
		GCN\_SVD             & 25.81         & 50.27          & 61.43          & 13.06         &  & 18.57         & 16.61          & 52.12         & 19.14         \\
		RGCN                 & 5.04          & 35.13          & 20.24          & 13.11         &  & 1.46          & 61.56          & 11.22         & 10.65         \\ \midrule
		ST-SparseGCN            & \textbf{4.33} & 48.42          & 17.44          & \textbf{7.21} &  & 1.92          & 60.74          & 17.96         & 4.73       \\
		ST-SparseGCN\_Jaccard   & 6.23          & \textbf{29.53} & \textbf{11.05} & 8.53          &  & \textbf{0.52} & 45.69          & \textbf{8.20} & \textbf{2.30} \\
		ST-SparseGCN\_SVD       & 24.87         & 47.02          & 59.13          & 13.40         &  & 18.22         & \textbf{15.82} & 47.82         & 19.31         \\
		\bottomrule
	\end{tabular}
	
	\label{tab:per}
\end{table*}

\subsection{Defense Performance Evaluation}
\label{sec:perturb}
In the section, we evaluate the overall defense performance of the proposed ST-SparseGCN by comparing it with various defense methods under different adversarial attackers along with different perturbation sizes. 

{\bf Perturbation Size.} For each attacker, we increase the perturbation rate from 0 to 0.25 with a step size of 0.05. In general, the defense performance decreases along with the increase of the perturbation size.
In order to concisely present the experiment results, we define a new metric to evaluate the defense performance, termed dropping rate (DR) as shown in Eq.(\ref{eq:DRmetric}).
\begin{equation}\label{eq:DRmetric}
DR(Acc,\widehat{Acc})=\frac{\widehat{Acc} - Acc}{Acc}
\end{equation}
where $\widehat{Acc}$ is the accuracy of GCN on clean/original graph. Dropping rate characterizes the defense performance by measuring the integration of the performance degeneration caused by attacker models and the performance remedy from the defense methods. The smaller the dropping rate, the better the defense methods. We use the mean dropping rate(mDR) to describe the overall defense performance along with different perturbation sizes.

{\bf Hybrid Defense.} To illustrate that the proposed ST-SparseGCN defense methods can be complementary to the existing defense methods, we propose to integrate ST-SparseGCN with two existing defense models, namely, GCN\_Jaccard and GCN\_SVD, which improve GCN robustness through data preprocessing. The two integrated defense models are called ST-SparseGCN\_Jaccard and ST-SparseGCN\_SVD, respectively.

\begin{figure*}[ht]
	\centering
	\includegraphics[width=\linewidth]{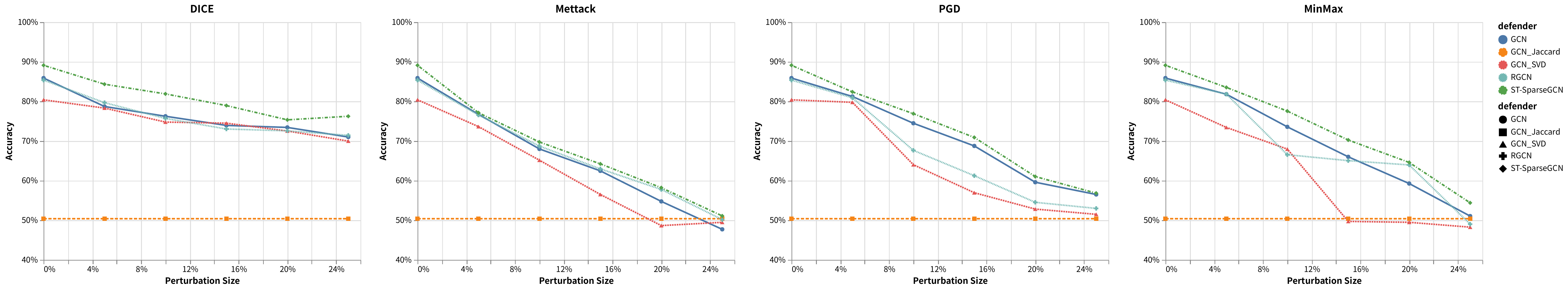}
	\caption{Results of different defenders when adopting different attackers in Polblogs datasets.}
	\label{fig:polblogs}
\end{figure*}

{\bf Experimental Results.} The experiment results on the Cora and Citeseer datasets are enumerated in Table~\ref{tab:per}. Due to space limitation, the experiment results on the Prolblogs dataset are not included, but illustrated in Fig.~\ref{fig:polblogs} instead.  

From Table~\ref{tab:per}, we can make the following observations: (i) the proposed ST-SparseGCN defense model or its variants (ST-SparseGCN\_Jaccard and ST-SparseGCN\_SVD) achieve the best defense performance under varoius attackers on all datasets, as ST-SparseGCN constructs a robust feature space in each GCN layer; (ii) the hybrid defenders (ST-SparseGCN\_Jaccard and ST-SparseGCN\_SVD) can improve defense performance compared with the corresponding baselines (namely GCN\_Jaccard and GCN\_SVD) alone in most cases, which implies that our proposed defender ST-SparseGCN is complementary to the existing  defenders, because ST-SparseGCN intends defend the adversarial attacks from the perspective of sparsification in feature space, which is complementary to most of the existing defense methods; (iii) none of the existing non-hybrid graph defenders (including the ST-SparseGCN alone) can perform best under all attackers on all datasets. This phenomenon may originate from the fact that the success of adversarial attacks comes from various aspects of the GCN model. Thus, the hybrid defense models may deserve further exploration.

Fig.~\ref{fig:polblogs} summaries the performance results under different attackers along with varied perturbation sizes on the Polblogs dataset. The results show that ST-SparseGCN again consistently achieves better performance than all the baselines, which demonstrates the superiority of the proposed ST-SparseGCN model. The experiment results shown in Table~\ref{tab:per} and Fig.~\ref{fig:polblogs} have illustrated that our defender can improve defense performance under all attackers on all datasets. It is worth to note that ST-SparseGCN does not rely on any prior knowledge of any particular adversarial attack method. The advantage of ST-Sparse might lie in that it can construct a robust feature space, which can be effectively against various adversarial attacks.

\subsection{SparseGCN and Dropout}
In the section, we compare the generalization performance and robustness of ST-Sparse and Dropout through experiments. 
Table~\ref{tab:dropout} demonstrates both dropout and ST-Sparse can improve the generalization ability of the model, and ST-Sparse is even better. In terms of robustness, ST-Sparse performs better than Dropout in the face of attacker. 
In addition, Dropout combined with ST-Sparse will damage the performance of the model. This phenomenon may be the random inactivation of Dropout will damage the ability to preferentially select features in ST-Sparse.

\begin{table}[]
	\caption{Defense performance(in percent) in classification accuracy with Dropout and ST-Sparse.}
	\centering
	\begin{threeparttable}
		\begin{tabular}{lll}
			\toprule
			& GCN          & ST-SparseGCN \\
			\midrule
			Clean        & 81.5$\pm$0.6 & 82.7$\pm$0.6 \\
			+Dropout     & 81.7$\pm$0.7 & 81.5$\pm$0.6 \\
			+Attacker         & 65.6$\pm$0.9 & 69.6$\pm$0.6 \\
			+Dropout+Attacker & 66.9$\pm$0.7 & 68.3$\pm$0.8 \\
			\bottomrule
		\end{tabular}
	 	\begin{tablenotes}
			\footnotesize
			\item[1] The dataset is Cora. The attacker is Mettack. The perturbation size is 0.05.
			\item[2] The results are averaged five times.
		\end{tablenotes}
	\end{threeparttable}
	\label{tab:dropout}
\end{table}

\subsection{Time Complexity}
\begin{figure}[ht]
	\centering
	\includegraphics[width=\linewidth]{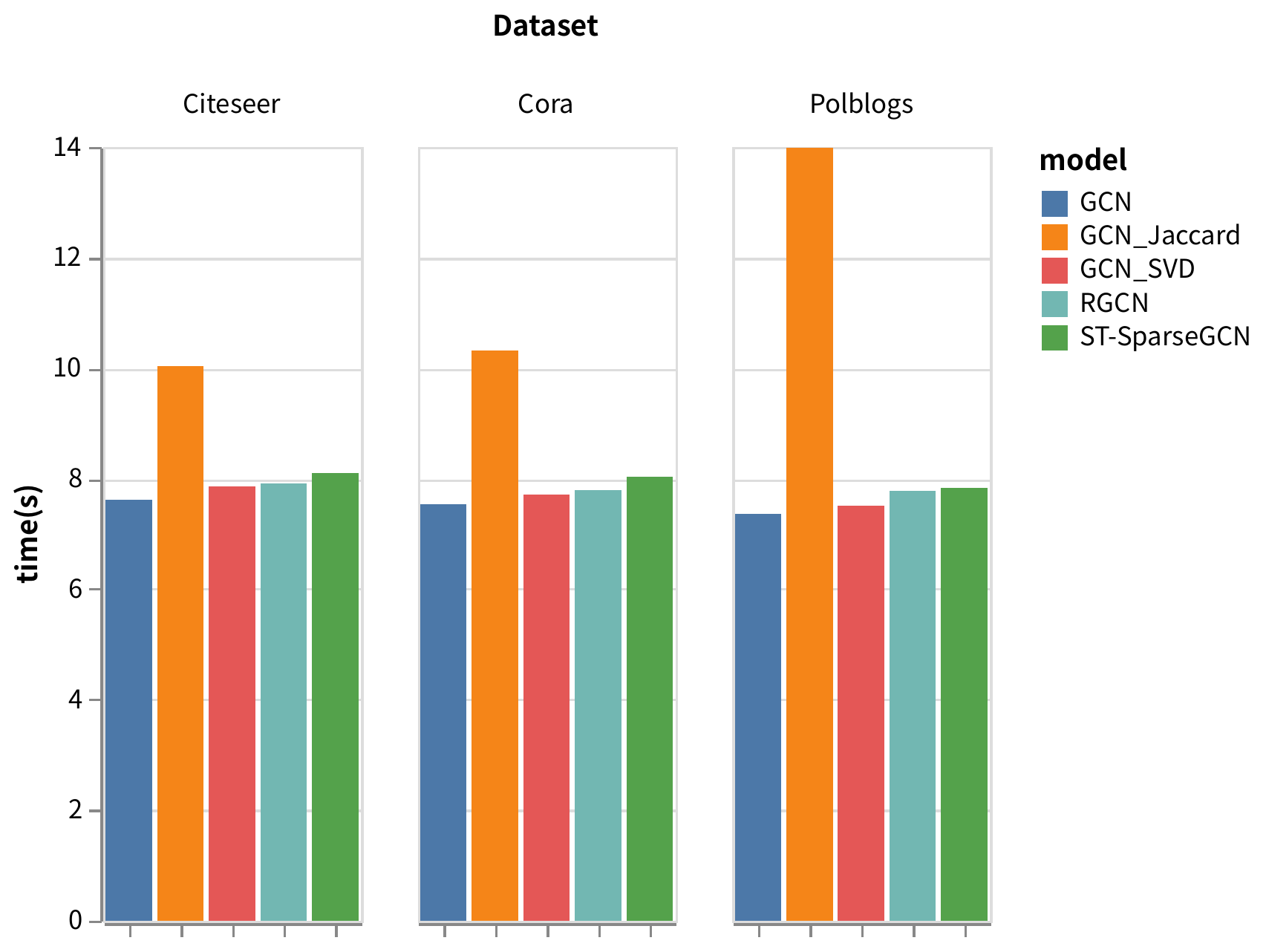}
	\caption{The time it takes to run the model once.}
	\label{fig:complex}
\end{figure}

We conduct several experiments on datasets-model pairs mentioned above to report the runtime of a whole training procedure for 200 epochs obtained on a single NVIDIA RTX 2080 Ti (cf. Fig.~\ref{fig:complex}). Thanks to the ability to quickly process sparse data in the PYG framework, the time overhead between models is basically the same. Among them, GCN\_Jaccard takes the most time because its data preprocessing process is very slow.

\subsection{Ablation Study and Parameter Analysis}
\label{sec:para}
In this section, we evaluate the maginal effect of the temporal sparsity, spatial sparsity, and the dimension $d_h$ on the accuracy and robustness of ST-SparseGCN. The performance evaluated on clean and perturbed datasets are shown separately. Due to space limitation, we only show the experiment results on the Cora dataset under the Mettack with a perturbation size of 0.05. The experiments on the other datasets exhibit similar patterns, which are included in the supplementary material. 

\begin{figure}[]
	\centering
	\subfigure[ST-sparsity]{
		\includegraphics[width=0.45\linewidth]{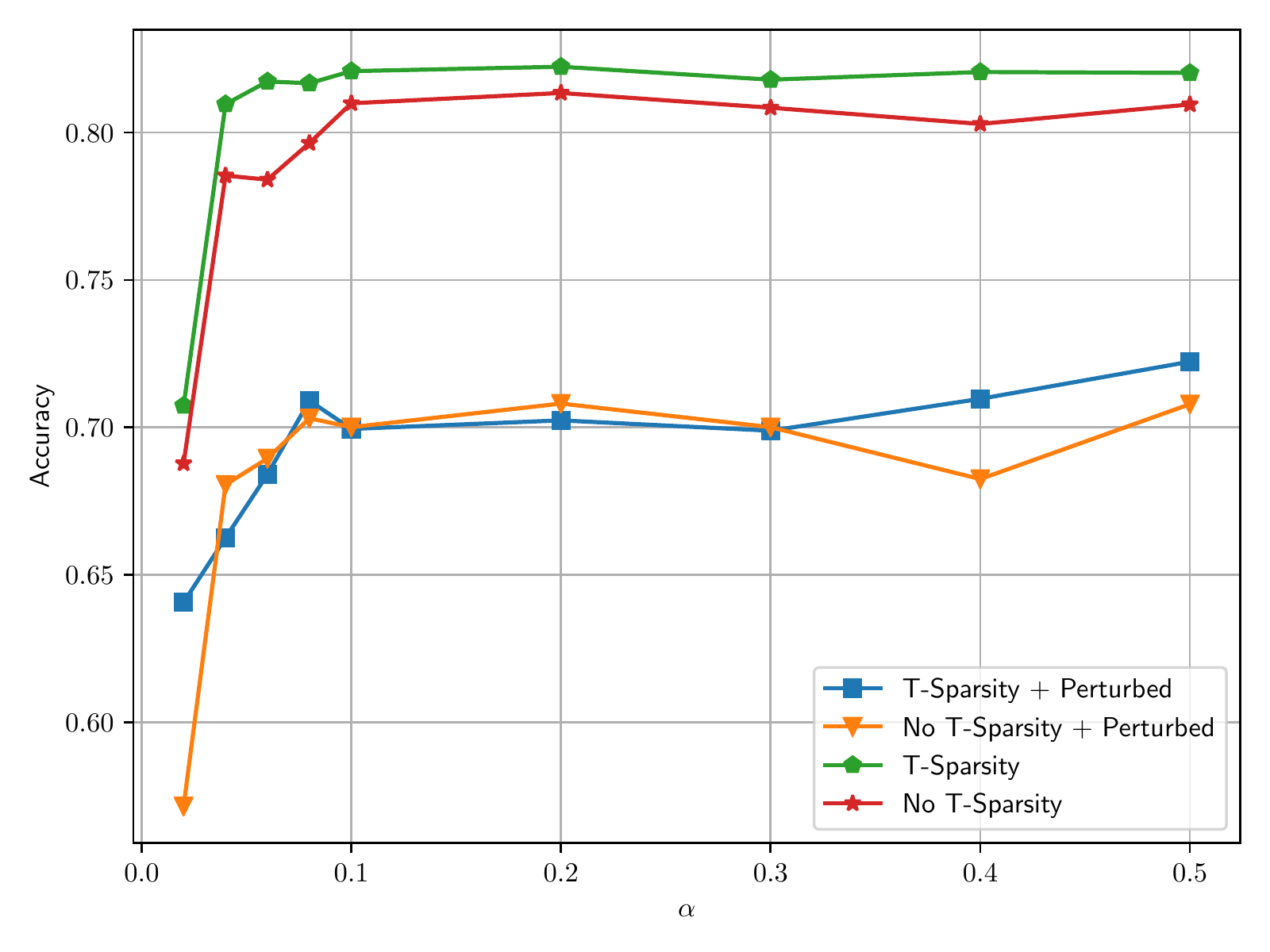}
		\label{fig:para:a}
		
	}
	\subfigure[Dimensions]{
		\includegraphics[width=0.45\linewidth]{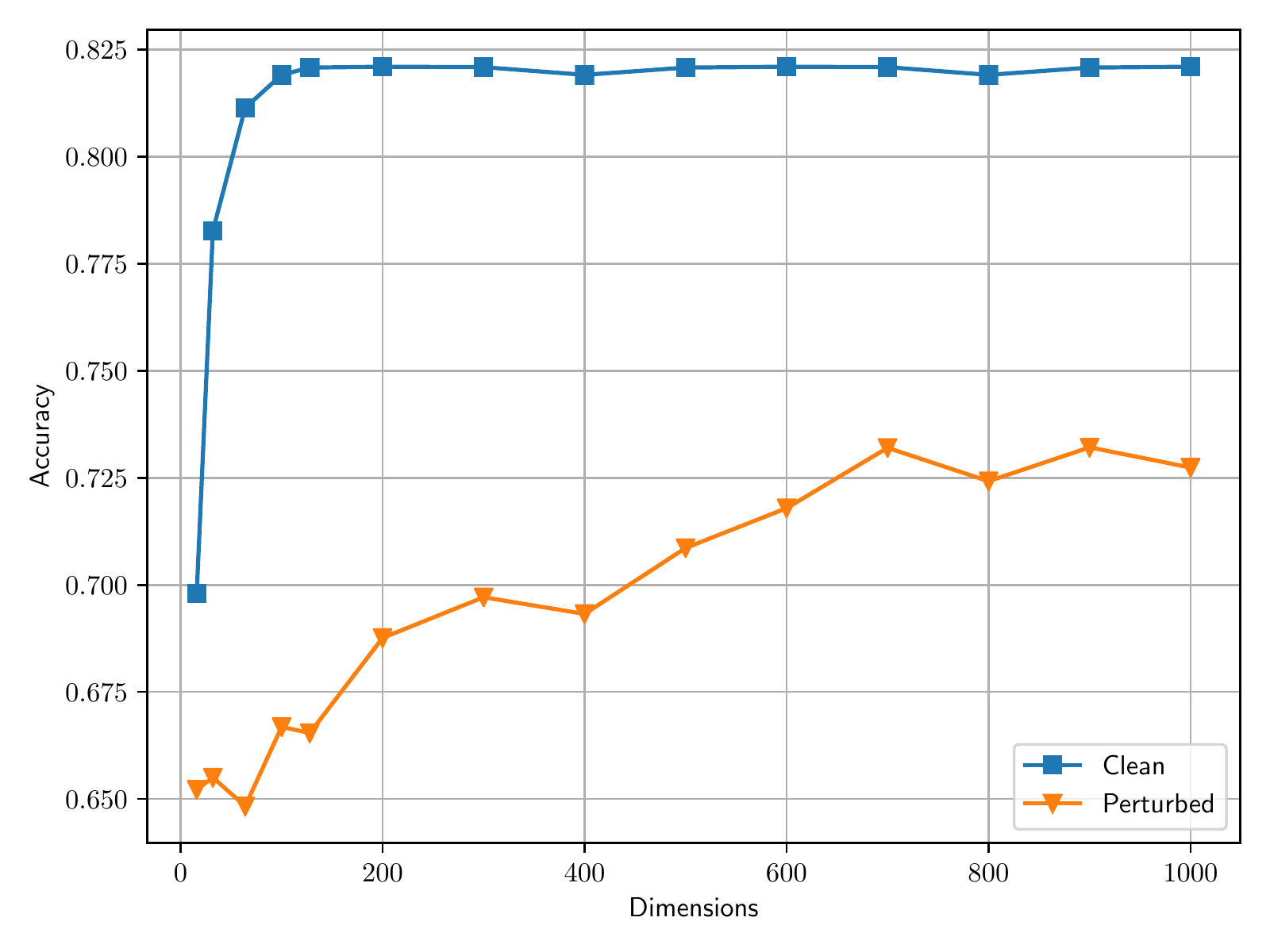}
		\label{fig:para:b}
	}
	\caption{Results of parameter analysis}
	\label{fig:para}
\end{figure}

%\subsection{Analysis of Spatio-Temporal Sparsity}
In the experiments, the extent of the spatial sparsity is controlled by the spatial sparse ratio $\alpha$. Fig.~\ref{fig:para:a} shows that the influence of the temporal sparsity and the Mettack along with the increasing sparse ratio $\alpha$ from $0.02$ to $0.5$, where T-Sparsity and Perturbed represent the temporal sparsity and the perturbation from the Mettack, respectively. From Fig.~\ref{fig:para:a}, by comparing the performance of the models with and without temporal spasity, it can be osberved that temporal sparsity can effectively improve the ST-SparseGCN's classification performance on both the clean graphs and the perturbed graphs. This illustrates the benefits of the temporal sparsity, which can not only increase the model's generalization capability (from the performance improvement on the clean graphs), but also improve the model's defense performance (from the performance improvement on the perturbed graphs).

It can also be inferred from Fig.~\ref{fig:para:a} that, when the spatial sparsity ratio $\alpha$ varies from a small value (0.02) to a relative larger value (0.08), both the models with and without temporal sparsity show significant performance improvement. Thus, this illustrates the necessity of spatial sparsity. Moreover, when $\alpha$ is larger than 0.08, the accuracy of the ST-SparseGCN remains basically unchanged. However, a very small $\alpha$ will degrade the performance, probably because the number of non-zero features is not enough to distinguish different categories for node classificaiton task.

\iffalse
Moreover, in different spatial sparsity, temporal sparsity will not harm the classification performance in the perturbed graph. An interesting question is why there is a difference in classification performance(improve and maintain) between clean graphs and disturbance graphs in temporal sparsity. This result may due to only using spatial sparsity may cause the model to fall into a local optimum, and this local optimum is robust.
\fi

Fig.~\ref{fig:para:b} illustrates the impact of the dimension $d_h$ on the performance. It is can be observed that whether it is the clean dataset or the perturbed dataset, the performance is drastically reduced when $d_h$ reduced to a small value. On the other hand, when $d_h$ increases to a certain level, the performance almost remains stable. This illustrates that there exists an appropriate value for $d_h$. The results also illustrate that the high-dimensional space can enable the GCN model to have more robust performance in case of the perturbation incurred by the attackers.

\section{Conclusion}\label{sec:conclude}
Although the GNN models have emerged rapidly, they still suffer the adversarial attack problem. Unlike the current works, which defend the attack on certain specific scenarios, this paper intends to universally address the attack problem. The proposed ST-Sparse mechanism is similar to the Dropout regularization technique in spirit, as it can provide a general adversarial defense layer, which can be readily integrated into numerous GNN variants. Meanwhile, ST-Sparse can also ensure both robust generalization and ordinary generalization. To evaluate the ST-Sparse's effectiveness, we conduct intensive experiments. The experiment results show that, in the face of four representative attack methods on three representative datasets with different levels of perturbation, ST-SparseGCN outperforms three representative defense methods.

\iffalse
\section{Acknowledgments}
This work was partially supported by the National Science Foundation of China, project nos.
61232001,61173169, 41871302, 91646115, 41571397 and 60903222; the Science Foundation of Hunan, project
no. 2016JJ2149 and no.018JJ3012; and the Major Science and Technology Research Program for Strategic Emerging
Industry of Hunan, grant no. 2012GK4054.
\fi

\bibliography{Lu.bib}
\end{document}